\documentclass[pmlr]{jmlr}

\usepackage{longtable}

\usepackage{enumitem}
\usepackage{titlesec}
\usepackage[skip=10pt]{caption} 
\usepackage{float}

\usepackage{booktabs}
\usepackage[load-configurations=version-1]{siunitx} 

\theorembodyfont{\upshape}
\theoremheaderfont{\scshape}
\theorempostheader{:}
\theoremsep{\newline}

\jmlrvolume{273}
\jmlryear{2025}
\jmlrworkshop{iRAISE 2025}

\title[Extended Reading Articles Generation with LLMs]{Stay Hungry, Stay Foolish: \linebreak On the Extended Reading Articles Generation with LLMs}

 \author{\Name{Yow-Fu Liou} \Email{alexliou.cs10@nycu.edu.tw}\\
  \Name{Yu-Chien Tang} \Email{tommytyc.cs10@nycu.edu.tw}\\
  \Name{An-Zi Yen} \Email{azyen@cs.nycu.edu.tw}\\
  \addr National Yang Ming Chiao Tung University, Hsinchu, Taiwan}

\begin{document}

\maketitle

\begin{abstract}

\label{sec:abstract}

The process of creating educational materials is both time-consuming and demanding for educators. 
This research explores the potential of Large Language Models (LLMs) to streamline this task by automating the generation of extended reading materials and relevant course suggestions. 
Using the TED-Ed \emph{Dig Deeper} sections as an initial exploration, we investigate how supplementary articles can be enriched with contextual knowledge and connected to additional learning resources.
Our method begins by generating extended articles from video transcripts, leveraging LLMs to include historical insights, cultural examples, and illustrative anecdotes. 
A recommendation system employing semantic similarity ranking identifies related courses, followed by an LLM-based refinement process to enhance relevance. 
The final articles are tailored to seamlessly integrate these recommendations, ensuring they remain cohesive and informative.
Experimental evaluations demonstrate that our model produces high-quality content and accurate course suggestions, assessed through metrics such as Hit Rate, semantic similarity, and coherence. 
Our experimental analysis highlight the nuanced differences between the generated and existing materials, underscoring the model's capacity to offer more engaging and accessible learning experiences.
This study showcases how LLMs can bridge the gap between core content and supplementary learning, providing students with additional recommended resources while also assisting teachers in designing educational materials.

\end{abstract}

\section{Introduction}
\label{sec:intro}

Crafting pedagogical materials is time-consuming and labor-intensive for educators, especially when striving to create engaging and comprehensive learning experiences. 
With the increasing advancements in LLMs, we see an opportunity to alleviate this burden by leveraging their capabilities in educational content creation. 
Recent studies have highlighted the potential of LLMs in transforming the educational landscape. 
For instance, researchers have demonstrated the use of LLMs in generating teaching materials, personalizing learning paths, and enhancing student engagement \citep{r1}. 
Additionally, LLM-driven intelligent tutoring systems have shown significant improvements in student satisfaction and learning outcomes by providing real-time feedback and personalized guidance \citep{r2}. 
Another study explored the customization of learning experiences using LLMs, showcasing their ability to adapt content to diverse learning styles and improve efficiency \citep{r3}. 
These works underscore the growing interest and progress in employing LLMs for educational innovation.

Building on recent advancements, this study explores the potential of LLMs to generate extended reading materials derived from initial articles, enhancing them with appropriate links to supplementary information to  support deeper learning. 
Specifically, we utilize courses from TED-Ed\footnote{\url{https://ed.ted.com/}} to investigate the relationship between video content and extended articles. 
By analyzing their recommended extended courses, we have developed a model designed to bridge the gap between initial articles and extended content.

Our research focuses on understanding the connections between extended articles and their source material, while integrating recommendation systems to generate continuously extending articles and course recommendations. 
The ultimate objective is to empower educators by providing tools to organize teaching content more efficiently and enable students to access an endless stream of supplementary information during self-study. 
Through these efforts, we aim to enhance the accessibility and quality of educational resources, making learning more dynamic and tailored to individual interests.

In summary, we make the following contributions: 
\begin{enumerate}[nolistsep]
\item We propose a novel approach to leverage LLMs for generating extended reading articles and recommending relevant courses, specifically using TED-Ed lessons as the research subject.
\item We develop and integrate a recommendation system that analyzes video content and its corresponding \emph{Dig Deeper} sections to generate stylistically aligned extended articles and course recommendations.
\item We conduct quantitative analyses to uncover stylistic differences between generated and actual \emph{Dig Deeper} content, providing actionable insights for future refinement.
\end{enumerate}
\section{Related Work}  \label{sec:relatedwork}

\subsection{LLMs in Educational Applications}
The application of LLMs for education has opened new possibilities for personalized learning and teaching support.
The study proposed by \citet{r4} examines how tools like ChatGPT can be utilized in education, highlighting both opportunities and challenges.
Similarly,  \citet{r5} provide a comprehensive overview of LLM applications in education, discussing their capabilities in content creation and personalized learning.
\citet{r6} present a method for generating programming projects to improve educational resources, concentrating on programming education. 
Besides the influence on the educational field, our research leverages LLMs to generate enriched educational content and incorporates a recommendation system that suggests relevant courses, providing a more comprehensive and personalized learning experience.

\subsection{Recommendation System and Scoring Tasks}

In the realm of recommendation systems and scoring tasks, traditional user-based collaborative filtering methods have been extensively utilized in e-learning platforms to generate personalized content recommendations based on user behavior \citep{r7}. 
These approaches rely on historical user data. 
On the other hand, our research leverages LLMs to automatically generate extended reading materials enriched with contextual knowledge, thereby enhancing the depth and relevance of educational content.
The advent of LLMs has significantly influenced the development of recommendation systems, particularly in understanding user preferences and generating personalized suggestions \citep{r8,r9,r10}. 
These models excel in processing natural language, enabling more flexible and accurate recommendations.
Additionally, LLMs have demonstrated potential in automated scoring tasks, such as evaluating English composition essays \citep{r11,r12,r13}. 
These studies highlight the capability of LLMs to assess textual quality and provide immediate feedback. 
Our research extends these applications by using LLMs not only to generate high-quality educational content but also to integrate course recommendations, thereby offering educators efficient tools and learners tailored educational pathways.
\section{Dataset} \label{sec:dataset}

Our data source comes from the TED-Ed website, which offers many lessons providing videos and post-class exercises. 
For each lesson, an extended reading section called \emph{Dig Deeper} is included for further reading and learning. 
When providing extended discussions related to the video content, it also adds recommended links. 
The structured resources on TED-Ed provide insights into the extensiveness and relevance of learning materials. 
By analyzing video transcripts, \emph{Dig Deeper} articles, and recommended links, we examine how the extensiveness of content impacts learning experience.
We collect the transcripts of every lesson video on the website, as well as the \emph{Dig Deeper} articles and their recommended links. 
Inspired by the concepts proposed by \citet{r14} and \citet{r15} that standardizing article lengths can enhance performance when comparing the relevance between two articles, we summarized the transcripts into articles of uniform length, aiming to maximize efficiency in evaluating both relevance and extensiveness. 
It's worth noting that we only used the TED-Ed lessons that recommend other on-site lessons in the \emph{Dig Deeper} section, along with their corresponding \emph{Dig Deeper} content, as our dataset.
The completed dataset of TED-Ed lessons (2,930 in total) serves as our database, providing students with more recommended resources. 
Besides, it supports teachers for the design of educational materials.
\section{Methodology}

\begin{figure}[htbp]
    \centering
    \includegraphics[width=0.8\textwidth]{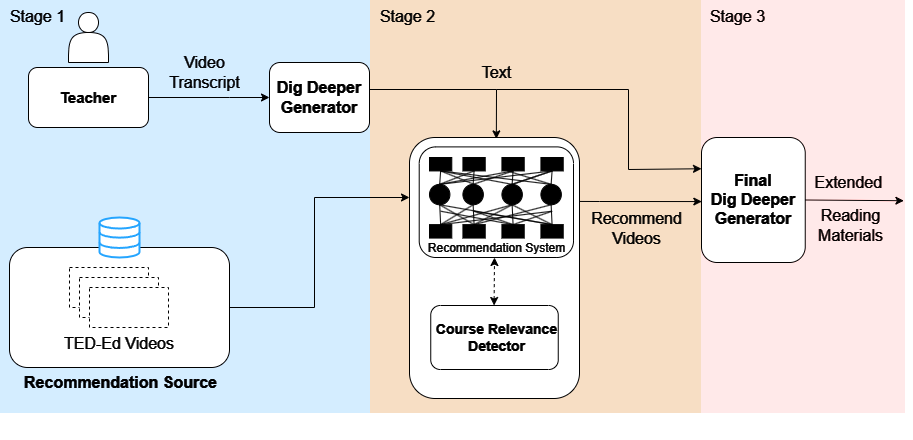} 
    \caption{Overview of System Framework.} 
    \label{fig:example} 
\end{figure}

In this section, we elaborate the approach used in our work. Figure~\ref{fig:example} presents an overview of our system framework. The framework consists of three main stages.
In Stage 1, We use an LLM to generate an initial \emph{Dig Deeper} article from the video transcript, incorporating relevant facts, terminology, and examples.
Then in Stage2, a sentence transformer and an LLM are used to generate recommendation content, while another LLM evaluates its reasonableness.
Finally in Stage 3, we refine the article by integrating selected lessons and keywords, enhancing its depth and connection to the recommendations.

\subsection{Stage 1: Generate the First Version of \emph{Dig Deeper}}
Based on our exploration of the TED-Ed website, we found that \emph{Dig Deeper} content often expands into areas such as historical facts, dates, events, terminology, and cultural practices, frequently providing concrete examples, case studies, or anecdotes to enrich the learning experience.
Therefore, we utilize an LLM as \emph{Dig Deeper} Generator to create an extended reading article centered on the transcript's main topic while incorporating these characteristics for each input video transcript. 
The generated article will serve as the initial version of the \emph{Dig Deeper} section.

\subsection{Stage 2: Recommend Relevant Lessons}
The generated article is compared with 2,930 lessons from our dataset using a sentence transformer to calculate similarity scores.
These scores are used to rank the lessons, and the top 100 lessons are selected as candidates for recommendation. 
These top 100 lessons, along with the generated article, are then input into an LLM-based recommendation ranking model. 
This model evaluates three key relationships between the generated article and each lesson: (1) Whether the lesson contains clearly related keywords from the article; 
(2) The overall relevance of the lesson to the article; 
(3) Whether the context of the keywords in the article aligns with the lesson’s content.

\subsection{Stage 3: Generate the Final Dig Deeper}

With the selected lessons and their associated keywords, we identify the positions of these keywords within the article. 
These keyword positions act as justifications for recommending the lessons. 
The original article is then rewritten to enhance its association with the recommended lessons, ensuring it maintains its depth and the integrity of the transcript’s main content. 
The outcome is the final version of the \emph{Dig Deeper}.
\section{Experiments \& Analysis}

To ensure the completeness of the articles and the quality of the recommended courses, our research focuses on the generated articles from three perspectives: (1) the appropriateness of the recommended links, (2) the relevance of the article content to the lesson's  transcript, and (3) the structural quality of the generated articles. 
Table \ref{tab:model-performance}  summarizes the evaluation results of two LLM models, Llama-3.1-405b and Gemma-2-27b, across multiple metrics, including Hit Rate, BERTScore \citep{r16}, BM25, Cosine Similarity, and Coherence Score. 
For Llama-3.1-405b, we access it through the SambaNova API\footnote{\url{https://sambanova.ai/}}.
For Gemma-2-27b, we conduct the experiments on an NVIDIA RTX 4090.

\begin{table}[t]
\centering
\captionsetup{skip=5pt} 
\caption{Results of two LLMs. The highest results for each metric are indicated in \textbf{boldface}, while the second best are \underline{underlined}.}
\label{tab:model-performance}\setlength{\tabcolsep}{8pt} 
\renewcommand{\arraystretch}{1.2} 
\resizebox{\textwidth}{!}{%
\begin{tabular}{|l|c|c|c|c|c|}
\hline
\bfseries Model & \bfseries Hit Rate & \bfseries BERTScore & \bfseries BM25 & \bfseries Cosine Similarity & \bfseries Coherence Score (LLM) \\ \hline

Llama-3.1-405b & \textbf{0.320} & \textbf{0.642} & \textbf{2.923} & \textbf{0.476} & \textbf{8.469} \\ \hline
Gemma-2-27b    & \underline{0.276} & \underline{0.559} & \underline{2.092}  & \underline{0.398} & \underline{6.958} \\ \hline
\end{tabular}%
}
\end{table}

\subsection{Quantitative Results}


\noindent \textbf{Hit Rate Evaluation.} The appropriateness of the recommended links is evaluated by comparing the links generated by our model with the original links provided on the TED-Ed website to calculate the hit rate. 
In Stage 1, our model generates an initial \emph{Dig Deeper} article without prior access to the database and then makes recommendations in Stage 2, based on the generated content. 
\begin{table}[t]
\centering
\captionsetup{skip=5pt} 
\caption{Results of Ablative Experiments.}
\label{tab:module-analysis}\setlength{\tabcolsep}{8pt} 
\renewcommand{\arraystretch}{1.2} 
\resizebox{\textwidth}{!}{%
\begin{tabular}{|l|c|c|c|c|c|}
\hline
\bfseries Method & \bfseries Hit Rate & \bfseries BERTScore & \bfseries BM25 & \bfseries Cosine Similarity & \bfseries Coherence Score (LLM) \\ \hline
Ours & 0.320 & 0.642 & \underline{2.923} & \underline{0.476} & \bfseries 8.469 \\ \hline
w/o Dig Deeper Generator    & \bfseries 0.515 & \underline{0.660} & 2.531  & 0.435 & 8.031 \\ \hline

w/o Final Dig Deeper Generator & \underline{0.413} & \bfseries 0.667 & \bfseries 2.990  & \bfseries 0.488 & \underline{8.034} \\ \hline
\end{tabular}%
}
\end{table}

\noindent \textbf{Relevance of Article Content.} The relevance of the article content is scored using three metrics: BERTScore, BM25, and cosine similarity. 
Extended readings on the TED-Ed website are not always closely related to the main topic; instead, they aim to provide readers with broader perspectives through the \emph{Dig Deeper} section. 
Our model adheres to this principle by attempting to extend the content without deviating from the core subject matter.

\noindent \textbf{Structural Quality of Generated Articles.} Evaluating the structural quality of articles lacks a well-established metric in current NLP evaluation methods. 
To tackle this issue, we employ LLMs to score the coherence of the generated extended articles from 1-10. 
The rationale behind this evaluation is that the \emph{Dig Deeper} sections on TED-Ed vary significantly---some merely offer links, while others present thoughtfully written long-form articles. 
Our objective is to generate extended articles that not only recommend additional TED-Ed lessons for readers to explore but also present a well-structured narrative suitable for reader engagement and comprehension.

\noindent \textbf{Ablation Studies.} To investigate the effects of this generated extended articles, Table \ref{tab:module-analysis} 
shows the results of our ablative experiments.
We can see that when we removed the initial generated extended articles and directly used the transcript for recommendations, the hit rate increased significantly.
We hypothesize that this is because of the diversity brought by LLMs, as they're instructed to generate in an exploratory manner.
This also aligns with the fact that our method achieves the highest score in coherence, suggesting that the initial generated extended articles enhance the structural quality of the final outputs.
\subsection{Analysis}
\label{subsubsec:Analysis}

To further understand different types of \emph{Dig Deeper} articles, we classify them into three categories based on their structure and content: (1) No text content, only links (2) Mainly text content, few links (3) Paragraph discussions with provided links.

\noindent \textbf{Category 1: No Text Content, Only Links.} The first type of \emph{Dig Deeper} uses one or two introductory sentences to guide readers toward clicking on recommended links. 
However, it lacks a complete article structure, detailed descriptions for the links, and comparable text content aligned with the transcript. Consequently, this type receives relatively low scores across various tasks, particularly in the evaluation of coherence.
\begin{table}[t]
\centering
\scriptsize
\captionsetup{skip=5pt} 
\caption{Results of Categorized Content.}
\label{tab:example}
\setlength{\tabcolsep}{8pt} 
\renewcommand{\arraystretch}{1.2} 
\resizebox{\textwidth}{!}{%
\begin{tabular}{|l|c|c|c|c|c|}
\hline
\bfseries Method & \bfseries BERTScore & \bfseries BM25 & \bfseries Cosine Similarity & \bfseries Coherence Score (LLM) \\ \hline
Ours & \bfseries 0.665 & 1.966 & \bfseries 0.528 & \underline{8.483}  \\ \hline
Category 1   & 0.514 & 1.149 & 0.379 & 2.422 \\ \hline
Category 2   & \bfseries 0.665 & \underline{2.341} & \underline{0.518} & \bfseries 8.692 \\ \hline
Category 3   & 0.477 & \bfseries 2.583 & 0.304 & 7.953 \\ \hline
\end{tabular}%
}

\end{table}

\noindent \textbf{Category 2: Mainly Text Content, Few Links.} The second type of \emph{Dig Deeper} provides a complete article with a strong structure and rich content but includes few links to related resources. 
This type of \emph{Dig Deeper} delves deeply into extended topics. 
In our evaluation, it scores highly in both relevance and coherence. 
However, it did not fully achieves the goal of encouraging extended reading by sharing recommendation links.

\noindent \textbf{Category 3: Paragraph Discussions with Provided Links.} The third type of \emph{Dig Deeper} makes up the majority on the website, serving as the target template we manage to produce. 
This type divides and discusses topics according to the knowledge points mentioned in the transcripts, providing multiple extended course links within each paragraph, which may compromise overall coherence. 
Thus, its coherence score is slightly lower than that of the second type. 
Additionally, it does not score particularly high in relevance due to its focus on extended exploration.

\section{Conclusion}
\label{sec:conclusion}

This study initiates a pilot exploration in leveraging LLMs to perform course recommendation tasks and generate extended reading articles.
By using TED-Ed lessons as the research subject, we analyzed the relationship between video content and the \emph{Dig Deeper} extended reading sections, and developed a model that integrates recommendation systems to produce extended articles and recommend relevant courses. 
Our quantitative experiments and analyses provided deeper insights into the stylistic differences between the generated \emph{Dig Deeper} sections and the actual content on TED-Ed, offering valuable directions for future improvements.
Moving forward, we plan to further investigate a better evaluation method to the explorative extent of generated articles and recommended courses, as well as enhance the adaptability of the model to different subject areas. 

\acks{
This work was partially supported by the National Science and Technology Council, Taiwan, under grant NSTC 113-2222-E-A49-007-.
}

\bibliography{reference}

\end{document}